\documentclass[10pt,journal,compsoc]{IEEEtran}
\ifCLASSOPTIONcompsoc
  \usepackage[nocompress]{cite}
\else
  \usepackage{cite}
\fi

\usepackage{algorithm,algorithmic}
\usepackage{animate}
\usepackage{cuted}
\usepackage{capt-of}
\usepackage{color}
\usepackage{xcolor}
\usepackage{comment}
\usepackage{amsmath}
\usepackage{graphicx}
\usepackage{caption} 
\usepackage{microtype}                 
\usepackage{textcomp}                  
\usepackage{mathptmx}                  
\usepackage{times}                     
\usepackage{tabu}                      
\usepackage{booktabs}                  

  \usepackage{hyperref}

\ifCLASSINFOpdf
\else
\fi
\begin{document}
%
\title{Casual 6-DoF: free-viewpoint panorama\\using a handheld 360° camera}
%
%
%
%

\author{Rongsen Chen,
        Fang-Lue Zhang,~\IEEEmembership{Member,~IEEE,}
        Simon Finnie,
        Andrew Chalmers,~\IEEEmembership{Member,~IEEE,}
        Taehyun Rhee,~\IEEEmembership{Member,~IEEE,}
        
\IEEEcompsocitemizethanks{\IEEEcompsocthanksitem R. Chen, S. Finnie, A. Chalmers and T. Rhee are with Computational Media Innovation Centre, Victoria University of Wellington, New Zealand.
\protect\\
E-mail: \{rongsen.chen, simon.finnie, andrew.chalmers, taehyun.rhee\}@vuw.ac.nz.

\IEEEcompsocthanksitem F.-L Zhang is with School of Engineering and Computer Science, Victoria University of Wellington, New Zealand. 
E-mail: fanglue.zhang@vuw.ac.nz. \protect\\
T. Rhee and F.-L. Zhang are the corresponding authors.}
\thanks{Manuscript received April 19, 2005; revised August 26, 2015.}}

%
%

\markboth{Journal of \LaTeX\ Class Files,~Vol.~14, No.~8, August~2015}%
{Shell \MakeLowercase{\textit{et al.}}: Bare Demo of IEEEtran.cls for Computer Society Journals}
%



\IEEEtitleabstractindextext{%
\begin{abstract}
Six degrees-of-freedom (6-DoF) video provides telepresence by enabling users to move around in the captured scene with a wide field of regard. Compared to methods requiring sophisticated camera setups, the image-based rendering method based on photogrammetry can work with images captured with any poses, which is more suitable for casual users. However, existing image-based rendering methods are based on perspective images. When used to reconstruct 6-DoF views, it often requires capturing hundreds of images, making data capture a tedious and time-consuming process.
In contrast to traditional perspective images, 360\textdegree{} images capture the entire surrounding view in a single shot, thus, providing a faster capturing process for 6-DoF view reconstruction. This paper presents a novel method to provide 6-DoF experiences over a wide area using an unstructured collection of 360\textdegree{} panoramas captured by a conventional 360\textdegree{} camera. Our method consists of 360\textdegree{} data capturing, novel depth estimation to produce a high-quality spherical depth panorama, and high-fidelity free-viewpoint generation. We compared our method against state-of-the-art methods, using data captured in various environments. Our method shows better visual quality and robustness in the tested scenes. 
\end{abstract}

}

\maketitle

\IEEEdisplaynontitleabstractindextext

%
\IEEEpeerreviewmaketitle

\IEEEraisesectionheading{\section{Introduction}}
Recent advancements in Virtual and Mixed Reality (VR/MR) have led to a surge in popularity of 360\textdegree{} panoramic media. They are well suited for VR/MR due to their wide field-of-view (FoV), which provides complete rotational freedom. In its current form, however, 360\textdegree{} media has a lack of freedom for translational motion which can break the user's immersion ~\cite{thatte2018towards, serrano2019motion}. 

Recent research of 6 degrees-of-freedom (6-DoF) media has been able to generate motion parallax in accordance with user motion. However, current approaches, such as Facebook's manifold RED camera~\cite{pozo2019integrated}, Google's welcome to light field \cite{overbeck2018system} and layered meshes~\cite{broxton2020immersive}, require sophisticated camera setups that are additionally reliant on professional capturing devices. The motion parallax generated by these methods is constrained to a small area, meaning they have limitations to provide free-viewpoint navigation over moderate to large distances. 

\begin{figure*}[t!]
  \centering
  \includegraphics[width=\linewidth]{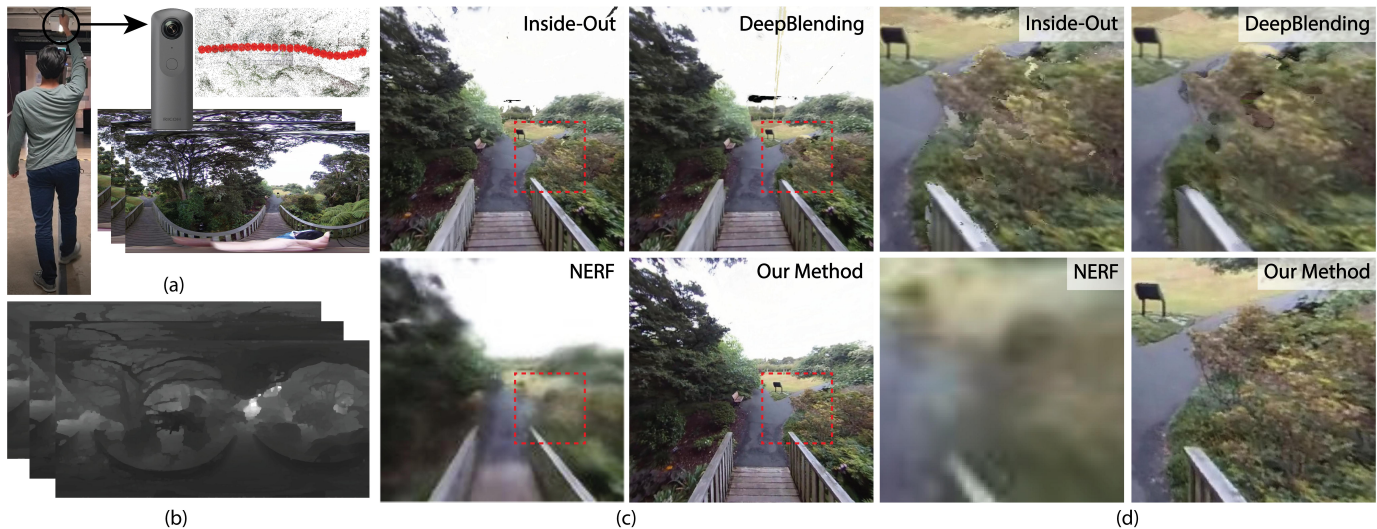}
  \caption{Our method provides 6-DoF experiences using 360\textdegree{} panoramic images captured by a handheld 360\textdegree{} camera. \textbf{(a)} The input 360\textdegree{} panoramas captured by a conventional 360\textdegree{} camera, \textbf{(b)} the corresponding depth panoramas estimated by our method, \textbf{(c)} synthesized novel views with comparisons, and \textbf{(d)} their corresponding zoomed in images.}
  \label{fig:teaser}
\end{figure*}

Image Base Rendering (IBR) methods~\cite{shum2000review, shum2008image} based on photogrammetry \cite{jancosek2011multi, schonberger2016structure}, on the other hand, only require an unstructured set of overlapping images that are more accessible for conventional (casual) users. These approaches can operate over larger areas due to their flexible camera setup and recovered geometric information. Inside-Out~\cite{hedman2016scalable} synthesizes high-quality free-viewpoint navigation using unstructured sets of RGBD images to enable 6-DoF. The method was further refined in DeepBlending~\cite{hedman2018deep}, where the captured depth maps were replaced with a depth estimation. These methods are still, however, limited by their use of perspective cameras, requiring large quantities of images to accurately capture full environments. This process is time-consuming, complex, and unpleasant for the casual user. 

The spherical 360\textdegree{} camera, in contrast to the perspective camera, has a complete view of the environment in each image (a perspective camera would normally take 5-8 shots to cover the same area). Using the 360\textdegree{} camera for the capturing process would greatly simplify the process for IBR-based 6-DoF methods. The use of 360\textdegree{} cameras in view synthesis has been explored recently in Omniphoto \cite{bertel2020omniphotos}. However, their approach was limited to generating views within a structured circle, restricting the user's range of motion while eliminating all vertical motion. In this paper, we overcome these limitations with the use of casually captured panoramas from a single 360\textdegree{} camera.

Our method provides real-time 6-DoF viewing experiences using 360\textdegree{} panoramic images captured by a handheld 360\textdegree{} camera (\autoref{fig:teaser}). Given an unstructured collection of 360\textdegree{} monocular panoramic images, our novel panoramic view synthesis method can synthesize panoramic images from novel viewpoints (a point in 3D space where no image was previously captured) in 30fps. Our method starts with an offline process to recover the orientation and position of each input panorama. We then recover the sparse and dense depth panoramas of the scene. Unlike previous methods~\cite{hedman2018deep, riegler2020free} that use this information to generate dense 3D geometric models for rendering new viewpoints, we directly synthesize 360\textdegree{} RGB images using the recovered depth from the input panoramas. We present a novel depth interpolation and refinement scheme that ensures high visual quality and fast view synthesis. 

We tested our method in various indoor and outdoor environments at medium to large scale scenes, with casually captured data using a consumer-grade 360\textdegree{} camera. We evaluated our method against current state-of-the-art approaches~\cite{hedman2016scalable, hedman2018deep, mildenhall2020nerf} in each of our environments over short and long ranges of motions.

Our contributions are summarized as follows:
\begin{itemize}
    \item We present a novel platform with a complete pipeline to enable 6-DoF viewing experiences using a set of panoramas captured by a handheld consumer-grade 360\textdegree{} camera.
    \item We present a robust approach to reconstruct the depth panoramas from a set of RGB 360\textdegree{} panoramic images. 
    \item We developed a novel method to synthesize novel panoramic viewpoints in real-time (30fps) to allow users to walk around within a large-scale captured scene.
\end{itemize}
\section{Related Work}


\subsection{6 Degrees-of-Freedom}
6-DoF methods have been attracting much attention in recent years due to the need for motion parallax in VR applications. Thatte et al.~\cite{thatte2017stacked} introduced stacked omnistereo, which uses two camera rigs stacked on top of each other in order to capture 6-DoF content. Welcome to Light Field~\cite{overbeck2018system} used a spinning camera setup for capturing high-density images and providing reliable depth estimation. Facebook Manifold RED~\cite{pozo2019integrated} is a camera system that has been specifically designed to capture 6-DoF video. More recently, Broxton et al.~\cite{broxton2020immersive} proposed a system that uses deep-learning to convert videos captured by a spherical camera rig into a layered mesh, which can be viewed in 6-DoF. These state-of-the-art capture methods were restricted to professional cameras and sophisticated setups. 
The photogrammetry-based methods such as depth image-based rendering (DIBR) \cite{zinger2010free, ndjiki2011depth, lipski2014correspondence} only require a set of overlapping photos which are not necessarily from the same camera, thus being more suitable for casually captured videos.

Inside-Out~\cite{hedman2016scalable} is a photogrammetry-based method that enables wide area, free-viewpoint rendering via tile selection. DeepBlending~\cite{hedman2018deep} uses a similar architecture to Inside-Out. One significant improvement of this method is that they use deep learning-based image blending to achieve higher visual fidelity. Recently, Xu et al.~\cite{xu2021scalable} further improved the work to have the ability to reconstruct reflection on a reflective surface. However, with the requirement of thousands of input images company with input depth from Kinect, thus, is not ideal for casual users such as tourists. Recent works such as NERF~\cite{mildenhall2020nerf, zhang2020nerf++}, free-view synthesis \cite{riegler2020free}, and stable view synthesis~\cite{riegler2020stable} used deep learning to render high fidelity novel views. However, these methods are slow and require significant running time to synthesize each frame. 

The methods discussed above were built for perspective cameras, meaning direct application to 360\textdegree{} panoramas would perform poorly. In this paper, we present a method for creating high fidelity 6-DoF scenes with 360\textdegree{} cameras.

\subsection{Panoramic 6 Degrees-of-Freedom}
Huang et al.~\cite{huang20176} presented an approach that used geometry estimated via Structure-from-Motion (SfM) to guide the vertex warping of a 360\textdegree{} video. This technique simulates the feel of a 6-DoF experience. However, the experience is somewhat lacking due to its inability to produce motion parallax. Cho et al \cite{cho2019novel} extended this method, allowing it to use multiple 360\textdegree{} panoramas as input. However, it still presents the same limitation. The lack of motion parallax, in either case, can lead to VR discomfort.

One of the early 6-DoF panorama applications which provided motion parallax was proposed by Serrano et al. \cite{serrano2019motion}. In their system, they used deep-learning to predict the depth map of a given 360\textdegree{} panorama and render a mesh representation of the environment (inherently allowing for motion parallax). They designed a three-layered representation to handle occlusion. However, their method only took input from a single viewpoint, meaning the output quality and range of motion were limited. MatryODShka~\cite{attal2020matryodshka} adapted Multi-Plane Image (MPI)~\cite{mildenhall2019local, zhou2018stereo} to 360\textdegree{} panoramas, creating a layered representation of Omnidirectional Stereo (ODS) images with deep learning, however, this approach also has a limitation on its synthesis quality and range of available motion.

In this paper, we present a novel method of performing wide-area free-viewpoint navigation using casually captured 360\textdegree{} panoramas.

\begin{figure*}
\centering
   \includegraphics[width=\linewidth]{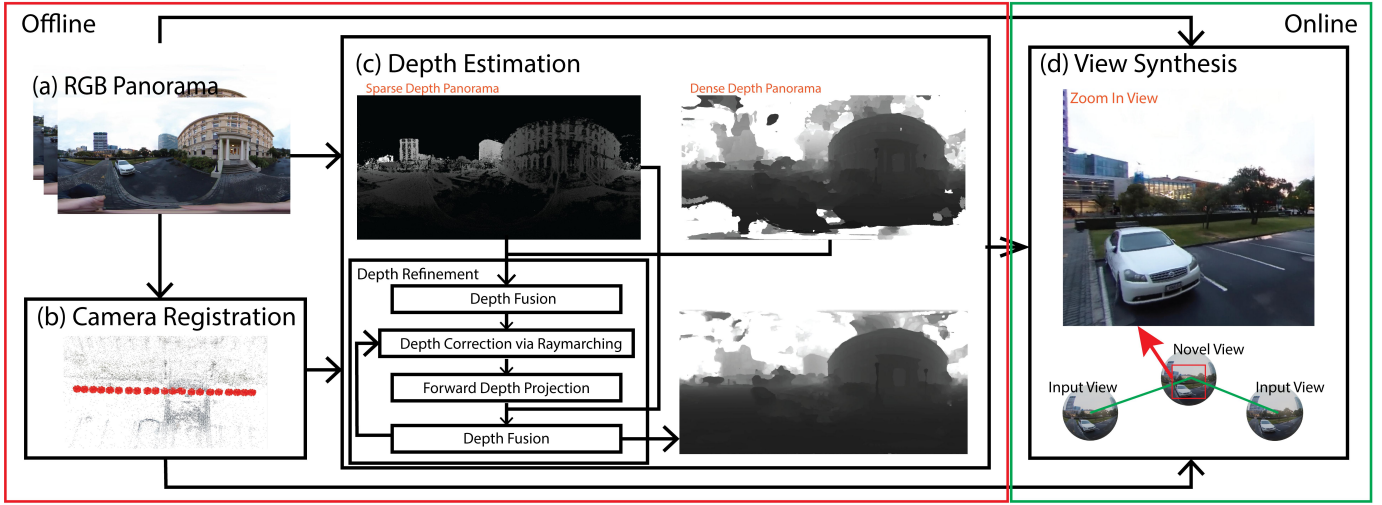}
\caption{The overview of our method. \textbf{(a)} The input 360\textdegree{} panoramas. \textbf{(b)} Camera parameter estimation. \textbf{(c)} Depth estimation  for the panoramas \textbf{(d)} View synthesize that enables high-fidelity 6-DoF viewing.}
\label{flowdiagram}
\end{figure*}

\subsection{Depth estimation}
Recovering depth information is important to enabling accurate 6-DoF images. Previous depth estimation methods for 360\textdegree{} images has attempted to recover the depth map via matched features \cite{sato2010efficient}, this type of method suffers from feature detection errors often producing an incomplete depth map. Most recent works on 360\textdegree{} images designed to perform depth estimation with single \cite{wang2020bifuse} or pair of 360\textdegree{} images\cite{zioulis2019, wang2020360sd}. These works estimate depth without validating depth consistency across images. The inconsistency across depth images causes severe ghosting artifacts and is not suitable for view synthesis of large scenes where dozens to hundreds of images are often used for reconstruction. To reconstruct consistent depth across multiple images, the depth needs to be estimated with the spirit of Multi-View Stereo (MVS).

Similar to other areas in computer vision, research has attempted to approach MVS using CNNs \cite{yao2018mvsnet, yao2019recurrent}. However, these methods often perform poorly on wildly captured images. Since their networks are trained by comparing the estimated depth with ground truth depth, it is limited to adapt the newly captured scenes when ground truth is not available. Neural radiance field (NeRF) \cite{mildenhall2020nerf} uses a Multi-Layer perceptron (MLP) to learn the presentation of the scene. Although this method requires training for every given new scene, the advantage is that it can train the network by directly comparing the estimated result with the capture RGB images, thus, more suitable for general scenes than methods trained with ground truth depth. However, we observed NeRF has difficulty in network converging when training with 360\textdegree{} panoramas. We demonstrate the limitation in \autoref{sec:RandD}. Since the learning-based methods were unable to provide acceptable depths from our survey and tests, we developed our depth estimation as in \autoref{sec:depth}.

\section{Overview}
An overview of our method is shown in Figure~\ref{flowdiagram}. It has an offline pre-processing phase and an online view synthesis phase. During the pre-processing phase, the captured 360\textdegree{} video sequence is first processed by registering camera parameters. Then the sparse and dense panoramic depths of each input panorama are estimated. In contrast to previous methods \cite{hedman2016scalable, hedman2018deep}, we estimated the dense depth map via a separate epipolor-based depth estimation instead of performing surface reconstruction on the sparse point cloud that builds using sparse depth maps. The estimated sparse and dense depths will then be refined to produce high-quality depth panoramas in several steps; sparse and dense depth fusion, depth correction via raymarching, and forward depth projection. The refinement process will reduce the noise in the estimated depths, as well as ensure depth consistency across input panoramas. 

The online phase is for novel 360\textdegree{} view synthesis, which can be done in real-time to support interactive VR applications. Given a target position to synthesize a novel view panorama, we firstly project all input panoramas to the target position based on their estimated depths, generating an initial RGBD 360\textdegree{} panorama. We generate dense depth panoramas by interpolating the depth values of the corresponding pixels from the input panoramas and further enhancing them with the same depth correction algorithm as used in the pre-processing phase. Using the enhanced depth panorama, we are able to synthesize the RGB values of the novel view panorama by retrieving RGB pixels from the neighboring input panoramas and using weighted blending to combine them.
\begin{figure*}
\centering
    \includegraphics[width=\linewidth]{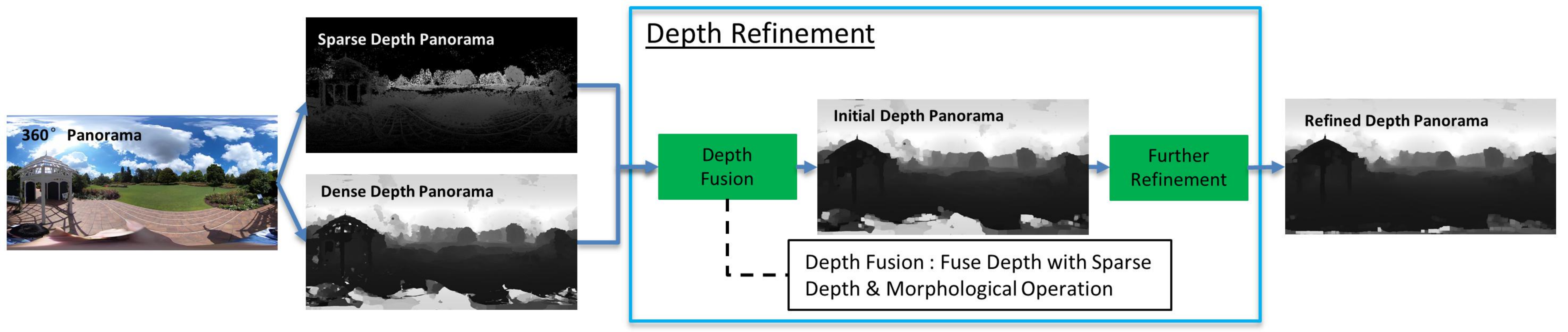}
\caption{The overview of our depth estimation pipeline. The process starts by generating depth panoramas from RGB inputs using the patch-based multi-view stereo and epipolar geometry. We then apply our depth refinement process on the fused depth panorama. \textbf{Depth Fusion} is consisting of depth panorama fusion in back-to-front fashion, and morphological operations. }
\label{depth_estimate}
\end{figure*}

\section{Capture and Registration}
\label{sec.4}
Our method takes a set of 360\textdegree{} panoramas as input, either discretely captured in different locations, or selected from a series frames in a video. Since blurring the image will harm the performance of camera registration methods (producing an inaccurate estimated camera pose), this will lead to ghosting artifacts in the view synthesis and degrade the quality of the estimated depth. When sampling frames from a video we use the variance of Laplacian Operator~\cite{bansal2016blur} to reject any blurred frames and substitute their closest sharp frames instead. Similar to other methods based on photogrammetry \cite{hedman2016scalable, hedman2018deep, riegler2020free, riegler2020stable}, our method made no assumptions regarding the camera transforms (translation and rotation). 

We recover the camera transforms corresponding to the selected frames using COLMAP (a SfM software package)~\cite{schonberger2016structure}. We chose COLMAP over other software such as Meshroom \cite{alicevision2021} because we found that COLMAP produces more robust registration result, as well as to ensure a fair comparison with the alternative methods ~\cite{hedman2018deep, mildenhall2020nerf} (which also used COLMAP).

However, COLMAP is designed for perspective images. In order to use it on 360\textdegree{} panoramas, we first project them into a set of perspective sub-images with overlapping FoV. Every panorama is projected into 8 perspective images, each with a 110\textdegree{} horizontal FoV, by ignoring the top and bottom poles of the 360\textdegree{} panorama. We sample perspective images in this way because we find it provides better input for COLMAP than cubemap projection in terms of feature quality and quantity. Since the 8 perspective images were sampled from the same 360\textdegree{} panorama, they shared the same center point, using this we can recover the position of the 360\textdegree{} camera.

\section{Depth Estimation for 360\textdegree{} Panorama
}\label{sec:depth}

Recent methods for view synthesis \cite{hedman2018deep, riegler2020stable} follow a process of first estimating sparse depths from RGB inputs, and then densifying them via surface reconstruction \cite{kazhdan2006poisson, cazals2006delaunay} to recover dense depth information for a given scene. However, since casually captured image-sets from non-professional users often lead to imperfect reconstruction, the reconstructed sparse depth will contain large regions of missing depths such as the complex outdoor scene as shown in \autoref{depth_estimate} (sparse depth panorama). It causes surface reconstruction errors, and thus, poorer results for view synthesis. To address this issue, we propose to use a two-stage approach to estimate the sparse depth and dense depth separately, using the patch-match depth estimation method and epipolar depth estimation method respectively. We further propose a depth refinement process to improve the quality of the estimated depth map. This refined depth will be later used for synthesizing novel views.

 

\subsection{Two Stage Panorama Depth Estimation}
\label{sec:two_statage_depth}

\noindent{\textbf{Sparse depth estimation:}} Our first step is to estimate a sparse depth panorama for each of the input RGB panoramas using the patch-match-based depth estimation method. We achieve this via COLMAP as in other similar methods\cite{hedman2018deep, riegler2020stable}. This approach produces good results for both small and thin objects, but it has limitations on texture-less regions, resulting in a depth panorama that contains a lot of missing information (referred to as a sparse depth panorama). Directly performing view synthesis with such depth would result in scenes with missing geometry. Thus, a denser depth information is required for a complete view synthesis result.
\\

\noindent{\textbf{Dense depth estimation:}} Previous methods \cite{hedman2018deep} obtain the dense presentation of the geometry via surface reconstruction using the sparse depth. However, the sparse depth estimated from casually image set may contains large gaps that is difficult for surface reconstruction to handle. Therefore, in contrast to previous methods, we propose to separately estimate the dense depth panorama using epipolar geometry~\cite{zhang1998determining}. Although this type of method often fails to detect thin geometries, it is able to recover more complete depth information of texture-less region than patch-base methods.

We implement the dense depth estimation as follows. Given a 360\textdegree{} panorama, we first generate sweep volumes directly using $N$ closest 360\textdegree{} panoramas (we use $N = 4$ in our experiment), given that it has a full FoV that could help to effectively avoid the out-of-FoV issue in narrow FoV videos. We then compare the sweep volume with the given 360\textdegree{} panorama to generate cost volume. We compute the cost volume using the classical ad-census \cite{mei2011building} method, because it is able to demonstrate good result on texture-less region. Similar to previous methods \cite{hosni2012fast, penner2017soft} we adapt guided image filters to filter the matching cost. Guided image filters smooth the filtered cost volume with regards to the edge of the guiding image, which helps sharpen the edge of the resulting depth images. We then follow the classical winner-takes-all depth selection to produce the dense depth map. 

\subsection{Depth Refinement}
\label{sec.depth-refine}
\begin{figure}
\centering
    \includegraphics[width=\linewidth]{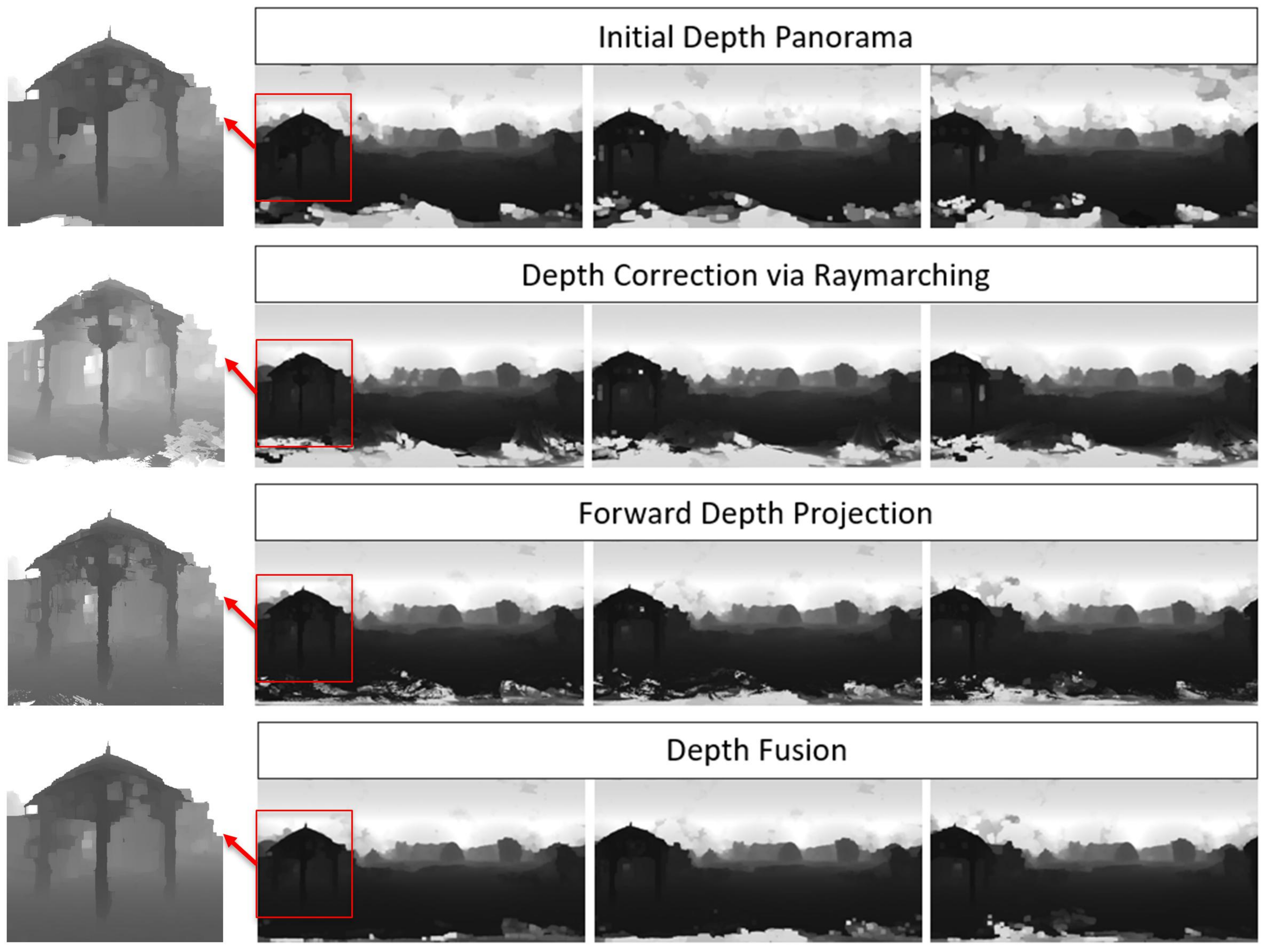}
\caption{The result from each refinement step. \textbf{Initial Depth:} Obtained by fusing the patch-matched based and epipolar based estimated depth together. \textbf{Depth Correction via Raymarching}: Correcting the depth that were too close to the camera using raymarching. \textbf{Forward Depth Projection:} Merging with neighboring depth via projection. \textbf{Depth Fusion:} use the result of forward depth projection as the new dense depth and combine it with the sparse depth map.} 
\label{stage_rs}
\end{figure}

Our dense depth panorama is estimated on a per-view basis, meaning that the estimated depth information is based on the parallax present in neighboring panoramas. Each panorama has different neighbors, meaning the recovered depth panoramas may not be consistent, as presented by the top row of \autoref{stage_rs}. This will cause visual artifacts during view synthesis. We solve the depth inconsistency, with an iterative depth refinement process. Our depth refinement process consist of three steps, depth fusion, depth correction via raymarching, and forward depth projection.
\\

\noindent{\textbf{Depth Fusion:}} Our depth fusion is a straight forward process that combines the dense depth map and sparse depth map in back to front fashion. During depth fusion we using the dense depth map as a hole filler to filling up the space where sparse depth estimation fails to reconstruct. Then we apply an morphological operation to the combined depth map to reduce the floating noise comes with the sparse depth. 

During the first iteration, we perform depth fusion using the estimated sparse and dense depth panorama. Similar to Hedman et al.~\cite{hedman2018deep}, our depth fusing allows us to adopt advantages of each depth type while avoiding their limitations. Nonetheless, the fused depth can still contain issues such as depth inconsistency that comes from the depth map estimated using epipolar approach. Thus, it needs further refinement to be used in view synthesis.
\\

\noindent{\textbf{Depth Correction via Raymarching:}} The depth consistency is optimized by finding the optimal depth value among neighboring panoramas. Floating geometry artifact often occurs when the estimated depth map has outliers that are too close to the camera. For instance, in \autoref{stage_rs} top row, the area around pavilion has estimated depth value much smaller than its neighboring area. We first attempt to remove those outlier points by checking weather an depth value is the highest depth that are agreed upon all neighboring depth panoramas in a given range. We check this consistency via a raymarching process, which is similar to the one used in Koniaris et al.~\cite{koniaris2018compressed} as shown in Algorithm \autoref{al1}. The increase rate $r$ and the nearest $K_{rm}$ input panoramas need to be set manually for different scenes in order to obtain optimal results. In our experiments, we set $r$ and $K_{rm}$ to 0.005 and 4 respectively by default, and found they worked well in all our scenes.

\begin{figure*}[!t]
\centering
      \includegraphics[width=\linewidth]{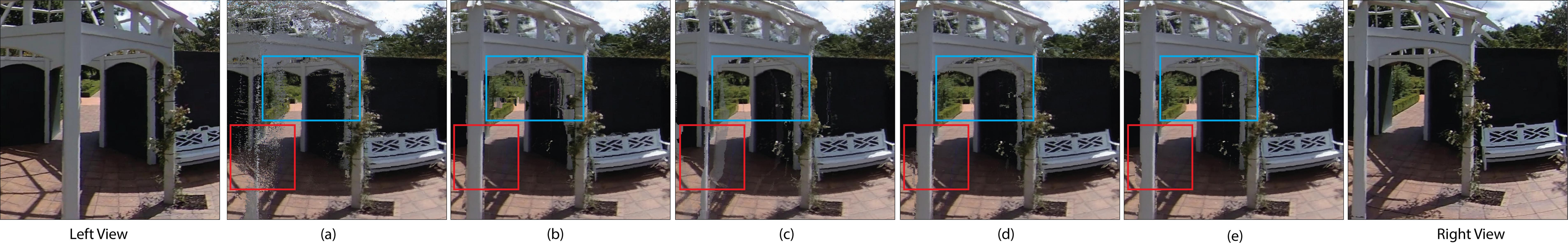}
\caption{View synthesis with each stage of the refinement result. The right most and left most are two closest reference views in the data-set. (a) Without refinement. (b) Result after applying depth fusion (initial depth). (c) Result after applying depth correction. (d) Result after forward depth projection. (e) Result after final depth fusion.}
\label{ablation}
\end{figure*}

 \begin{algorithm}
 \caption{Depth Correction via Raymarching Algorithm}
 \begin{algorithmic}[1]
 \renewcommand{\algorithmicrequire}{\textbf{Input:}}
 \renewcommand{\algorithmicensure}{\textbf{Output:}}
 \REQUIRE for any given depth panorama $D$, it's corresponding position $\mathbf{t_{new}}$, the number of nearest neighbors $K_{rm}$, the collection of neighboring input depth panoramas $\{D_1, D_2, \ldots, D_K\}$, their corresponding positions $\{\mathbf{t}_1, \mathbf{t}_2, \ldots, \mathbf{t}_K\}$, increase rate $r$.
 \ENSURE corrected depth $D_{cor}$
   \STATE $D_{cor} = zero(Width, Height)$
  \FOR {$j = 1$ to $Height$}
    \FOR {$i = 1$ to $Width$}
      \STATE $d = D(i, j)$
      \FOR {$k = 1$ to $K_{rm}$}
         \STATE {$i_{rep}, j_{rep}, d_{rep} =$ reprojection($i, j, d, \mathbf{t_{new}}, \mathbf{t}_k$)}
         \IF{$d_{rep} < D_k(i_{rep}, j_{rep})$}
           \STATE $d = d + rd$
           \STATE $k = 1$
         \ENDIF
      \ENDFOR
      \STATE $D_{cor}(i, j) = d$
    \ENDFOR
  \ENDFOR
 \RETURN $D_{cor}$ 
 \end{algorithmic} 
 \label{al1}
 \end{algorithm}

Although our raymarching process able to removes the  outlier depths that were too close to the camera, it often erased part of object. This is because, due to the limitation of epipolar based depth estimation, not all neighboring views would be correctly estimated for thin objects, if the sparse depth map fails to address, the raymarching process would end up removing the depth information of that object. An example has been shown in  the second row of \autoref{stage_rs}, where the pillar of the pavilion has been partially erased. In such cases, the refined depth panorama will lose the depth information related to the occluding object, causing artifacts where geometric details are missing. We overcome this issue via our following forward depth projection process.
\\

\noindent{\textbf{Forward Depth Projection:}} Our depth correction refines the depth map by checking the consistency of depth values across neighboring panoramas within a given range. Since each panorama has different neighbors, the result of the depth correction will be slightly different. Although an occluded region in one image might be partially erased during the first depth refinement process, such information may still exist in one of its neighboring panoramas. Our forward depth projection is able to make use of this view-dependent depth information, by merging a given depth and its neighboring panoramas to recover lost occlusion information. Our depth projection is shown in Algorithm \autoref{al2}. Where, the neighborhood range $K_{fp}$ is scene dependant. Then, we set the value $K_{fp} = K_{rm} + 2$, in our experiments. The slightly wider search range obtains information from the neighboring panoramas that are less likely affected by the same problematic depth panorama during raymarching.

\begin{algorithm}
\caption{Forward Depth Projection Algorithm}
\begin{algorithmic}[1]
\renewcommand{\algorithmicrequire}{\textbf{Input:}}
\renewcommand{\algorithmicensure}{\textbf{Output:}}
\REQUIRE position of the target viewpoint $\mathbf{t_{new}}$, the number of nearest neighbors $K_{fp}$, the collection of neighboring input depth panoramas $\{D_1, D_2, \ldots, D_K\}$, their corresponding positions $\{\mathbf{t}_1, \mathbf{t}_2, \ldots, \mathbf{t}_K\}$.
\ENSURE  fused depth $D_{fus}$
 \STATE $D_{fus} = infinity(Width, Height)$
 \FOR {$k = 1$ to $K_{fp}$}
  \FOR {$j = 1$ to $Height$}
    \FOR {$i = 1$ to $Width$}
         \STATE $d = D_k(i, j)$ 
         \STATE {$i_{rep}, j_{rep}, d_{rep} =$ reprojection($i, j, d, \mathbf{t}_k, \mathbf{t_{new}}$)}
         \IF{$d_{rep} < D_{fus}(i_{rep}, j_{rep})$}
            \STATE $D_{fus}(i_{rep}, j_{rep}) = d_{rep}$
         \ENDIF
      \ENDFOR
    \ENDFOR
 \ENDFOR
\RETURN $D_{fus}$ 
\end{algorithmic} 
\label{al2}
\end{algorithm}

For each iteration of depth refinement, we increase the value of $K_{rm}$ and $K_{fp}$ slightly (we use 2 in our experiment). The slightly increased checking range helps to improve the consistency of the depth over more panoramas, eventually $K_{rm}$ and $K_{fp}$ will be equals to the number of input panoramas. However, in our experiments, we found that around 3 iterations was sufficient for most of our test scenes. We illustrate each step of our depth refinement in \autoref{stage_rs}, while showing the result of each refinement step. We also illustrated a few synthesized views using the depth maps from each depth refinement step to demonstrate the quality improvement.

\section{Free-Viewpoint Synthesis}
\label{sec.Free-Viewpoint Rendering}
Our next step is to synthesize novel view panoramas using a set of input RGB panoramas and their corresponding depths and transformations obtained by our pre-processing phase. Previous methods \cite{hedman2016scalable, hedman2018deep} relying on the mesh reconstruction for each input image, leading to accumulated geometric errors when blending each mesh for novel view synthesis, thus limit the quality. To avoid this, we approach our view synthesis via depth based image warping with real-time depth correction. During view synthesis, we first generate the depth panorama for the synthesized view with depth correction from the input panoramas. Then, use this depth panorama to extract RGB pixel values from inputs, and weighted blend them to synthesis the pixel colors of the novel view panorama.

\subsection{Spherical 3D reprojection}
\label{sec:sd}
\textbf{Image to 3D coordinates: }We first aligned orientations of input panoramas to ensure they facing the same direction. Since we use the equirectangular representation of 360\textdegree{} panoramas as in the prior work \cite{thatte2017stacked}, we first convert the pixels $(i, j, d)$ of the input equirectangular RGBD images into spherical polar coordinates $(\theta, \phi, d)$, where $\theta \in [-\pi, \pi]$ and $\phi \in [\frac{-\pi}{2}, \frac{\pi}{2}]$ and $d$ is the corresponding depth value from input depth panorama $D_k$. We then project all the converted spherical coordinates of each input panorama into local 3D Cartesian coordinates $\vec{v}$ by \autoref{eq_1}.

\begin{equation}
\begin{bmatrix}
v_x\\
v_y\\
v_z
\end{bmatrix}
=
\begin{bmatrix}
d\cos{(\phi)}\cos{(\theta)}\\
d\sin{(\phi)}\\
-d\cos{(\phi)}\sin{(\theta)}
\end{bmatrix}
\label{eq_1}
\end{equation}
 
\textbf{Forming novel panorama: } The next step is to reproject $\vec{v}$ into the center of target novel view panorama in its local coordinates. If we define the center of target novel view panorama as $\mathbf{t_{new}}$, then the transformation between any other input panorama which centered at $\mathbf{t}_k$ can be calculated using $\vec{t} = \mathbf{t_{new}} - \mathbf{t_k}$. Then all 3D points $\vec{v}$ of input panoramas are reprojected into the target position $\mathbf{t}_{new}$ by $(\vec{v} - \vec{t})$. Then, the reprojected points of each input can subsequently be converted into spherical polar coordinates $(\theta_{rep}, \phi_{rep}, d_{rep})$ of $\mathbf{t}_{new}$ using \autoref{eq_2}. 

\begin{equation}
\begin{bmatrix}
\theta_{rep}\\
\phi_{rep}\\
d_{rep}
\end{bmatrix}
=
\begin{bmatrix}
\arctan{(-\frac{(v_z - t_z)}{(v_x - t_x)})}\\
\arcsin{(\frac{(v_y - t_y)}{d_{rep}})}\\
\lvert\lvert \vec{v} - \vec{t} \rvert\rvert
\end{bmatrix}
\label{eq_2}
\end{equation}

This spherical polar coordinates then converted into pixels of the equirectangular representation of the novel view panorama. 

We use Equation \ref{eq_1} and \ref{eq_2} to extract the necessary data from the input panoramas, and synthesize the novel views in two steps: 1) Backwards warping with depth correction, and 2) image blending, described as follows. 

\begin{figure*}
\centering
    \includegraphics[width=\linewidth]{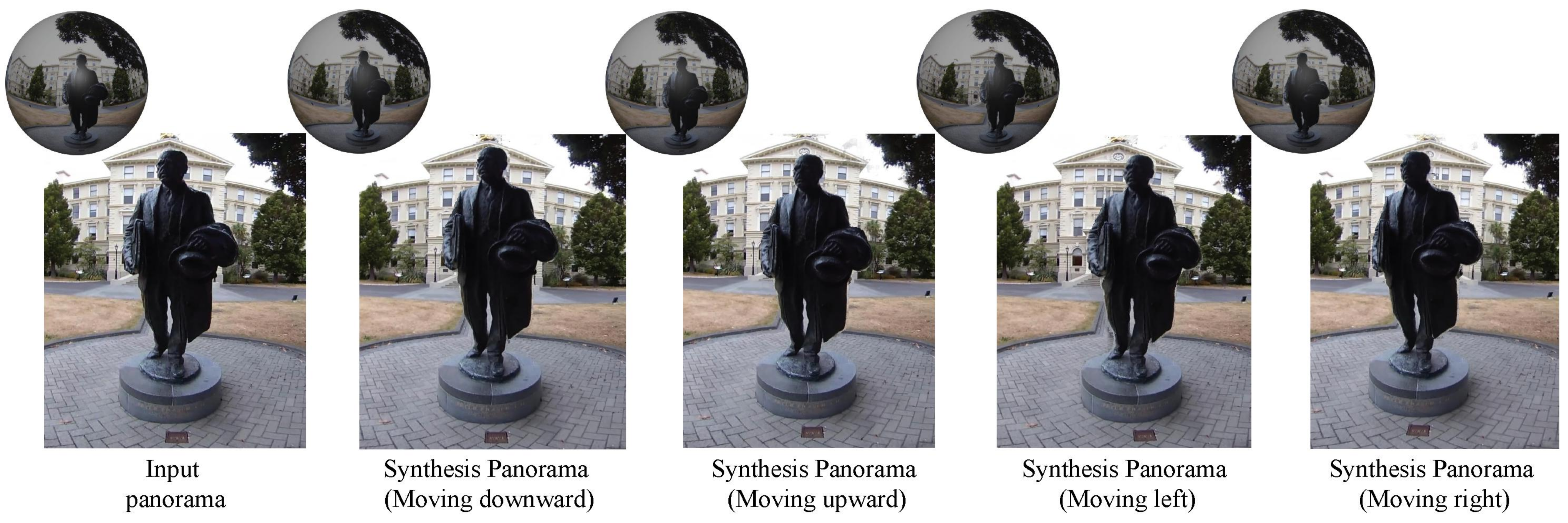}
\caption{Examples of novel view synthesis from various directions.}
\label{samplegeneratedview}
\end{figure*}

\begin{figure}
\centering
      \includegraphics[width=\columnwidth]{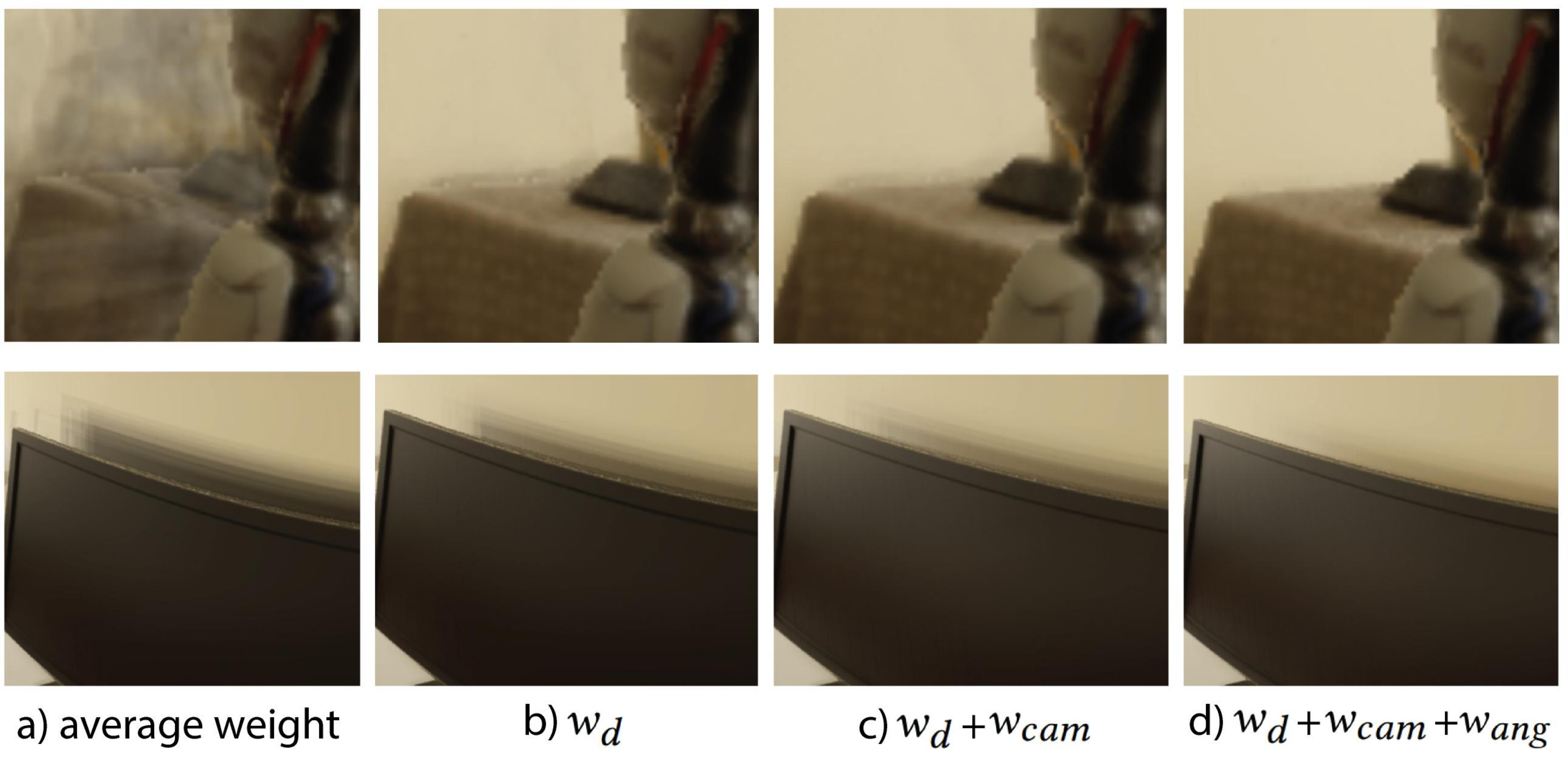}
\caption{Synthesized view using different strategies. \textbf{(a)} Novel view synthesized with the average weight, \textbf{(b)} blending with $w_d$ only, \textbf{(c)} blending with both $w_d$ and $w_{cam}$, and \textbf{(d)} is the result when our full weighting formulation applied.}
\label{fig3-5}
\end{figure}

\subsection{Backwards Warping with Depth Correction}
\label{sec:warping_refine}
Direct projection of the input panoramas into a novel view tends to produce many missing pixels. Simple color interpolation such as using linear or cubic function result in a blending of foreground and background.

To overcome this, we first synthesize a depth panorama at the target novel view, interpolate the depth values to avoid any missing values. In our approach we adapted a morphological closing \cite{koskinen1991soft} similar to the one used by thatte et al.~\cite{thatte2017stacked} in order to achieve the desired result. 
%
We use the synthesized depth panorama to extract RGB pixel values from each input panorama using Equation \ref{eq_1} and \ref{eq_2}, to synthesis the novel view panorama.  

However, there may be disagreement between the input depth panoramas for what the correct depth value is at the target position. A depth value that refers to the correct pixel in one input panorama may point to a different location in another input panorama, leading to visual artifacts in the synthesized panorama. To correct this, we apply Algorithm \autoref{al1} to the synthesized depth panorama, improving the overall visual consistency. We perform backwards warping with the corrected depth panorama in order to obtain the RGB pixels from the input panoramas with reduced visual artifacts.

\subsection{Image Blending}
\label{sec:image_blending}
To synthesis color values of our novel view panorama, we use the synthesized depth panorama to extract the RGB pixel values from the corresponding input panoramas. We extract the pixel values from the closest $K$ neighboring panoramas, where $K = 4$. If a suitable pixel cannot be found from closest neighbors, progress further to find the corresponding pixels across other input panoramas. 

After extracting corresponding pixel values, their RGB values are blended  by our weighting formula. Our weighting formula is based on three components: the correctness of the estimated depth $w_d$, the distance between the position of target view and input panorama $w_{cam}$, and the angle between the viewpoints and the relevant point in the scene $w_{ang}$. Then, the final blending weight $W$ for each pixel is computed by:
\begin{equation}
W = w_dw_{ang}w_{cam}
\label{eq_6}
\end{equation}
\\

The $w_d$ is computed by the difference between the estimated depth and the actual depth value as:
\begin{equation}
w_d = (|d_{rep} - d| + 1)^{-1}
\label{eq_4}
\end{equation}
where $d_{rep}$ is the estimated depth by reprojection from the position of the novel view depth panorama to the position of input depth panorama, and $d$ is a depth value at the input depth panorama. If the difference between $d$ and $d_{rep}$ is large, it means an occlusion occurs, and the pixel from this input panorama should contribute less towards our output. 

The camera weight $w_{cam}$ is used to measure the distance between the center of the novel view panorama and the center of the input panorama defined  as:
\begin{equation}
w_{cam} = \frac{s}{\lvert\lvert \vec{t}\rvert\rvert}
\label{eq_6}
\end{equation}
where $\vec{t}$ is described in \autoref{sec:sd}. The constant $s$ is the scale weight, and we set it as $10$ in our experiment.

Our angle weight $w_{ang}$ is to reproduce view dependent features in the synthesized image by favoring pixels with similar view angles as considered in \cite{buehler2001unstructured, bertel2020omniphotos}. The $w_{ang}$ is calculated by the angle between the vector $\vec{t}$ and vector $\vec{v}$ , defined as:
\begin{equation}
w_{ang} = \pi - \arccos{\left(\frac{\vec{t}\cdot\vec{v}}{\lvert\lvert \vec{t} \rvert\rvert \lvert\lvert \vec{v} \rvert\rvert}\right)}
\label{eq_5}
\end{equation}

Finally, each pixel value of the novel panorama $I_n(p)$ is synthesised by weighted blending of corresponding pixel values from $K$  neighboring input panoramas $I_{k}$ as:
\begin{equation}
I(p) = \frac{\sum_{k=1}^K W_{k}I_{k}(q_k)}{\sum_{k=1}^K W_{k}}
\label{eq_6}
\end{equation}

where $q_k$ represents the corresponding pixel in the $k$-th input panorama. We illustrate the effect of different weighting formulation in \autoref{fig3-5}. Our weighted blending successfully synthesis pixel values of the novel view panorama while reducing blending errors such as ghosting artifacts. 
\begin{figure*}
\centering
    \includegraphics[width=\linewidth]{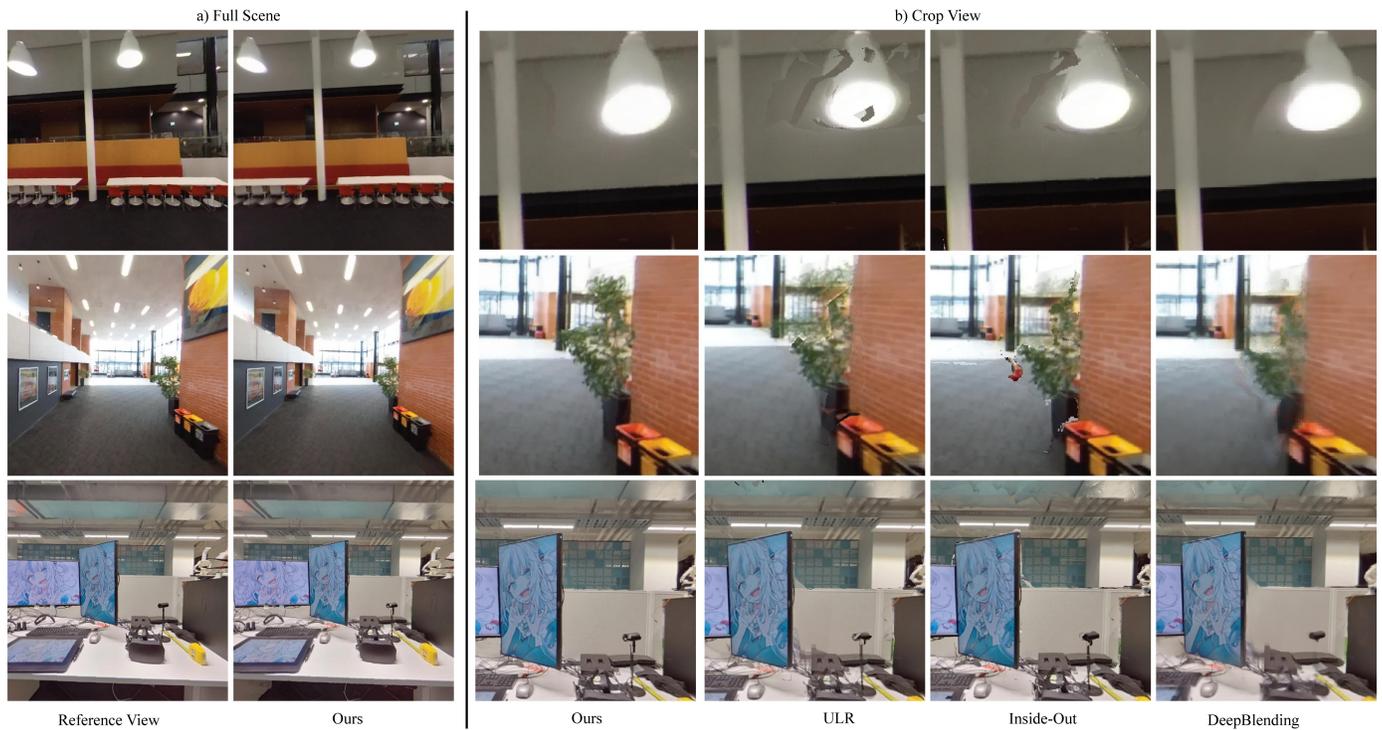}
\caption{Test results for the three indoor scenes. ULR fails in all cases. Compared to Inside-Out our method shows fewer visual artifacts on the borders and surfaces of objects. Unlike DeepBlending, our method looks sharper with more texture detail.}
\label{qualitative_in}
\end{figure*}

\begin{figure*}
\centering
    \includegraphics[width=\linewidth]{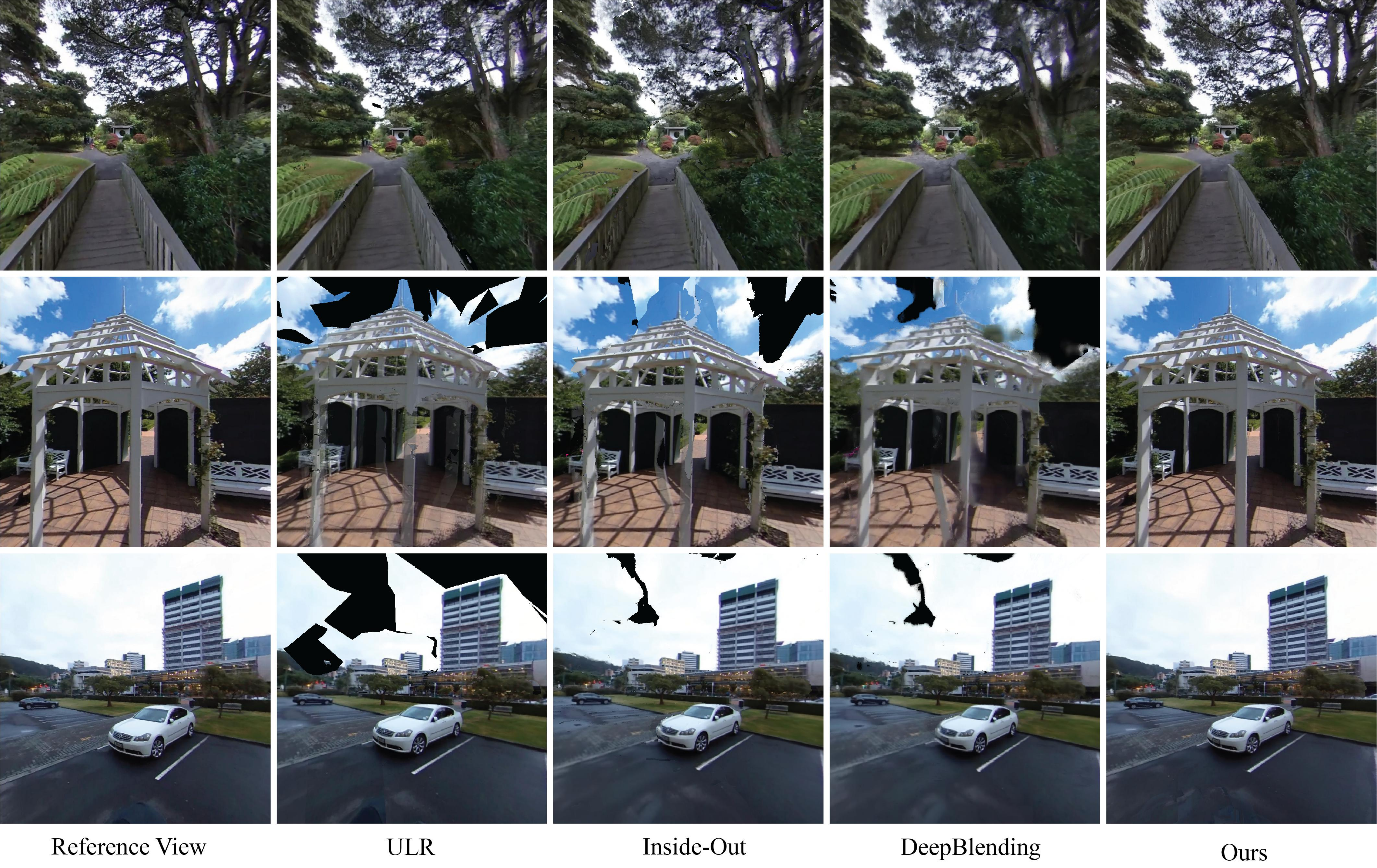}
\caption{Results of the three outdoor scenes. Our method presents more robust view synthesis for large outdoor scenes in situations where others failed.}
\label{qualitative_out}
\end{figure*}

\section{Data and Implementation}
We used the Nvidia CUDA library for GPU-based computations such as view synthesis. We rendered our novel view panorama to a VR HMD using the Oculus SDK and OpenGL libraries. The current implementation is able to work at 30fps.

We evaluated our method with our own captured datasets for the following two reasons. First, there are not any available benchmark datasets used for evaluating free-viewpoint synthesis methods that consist of 360\textdegree{} panoramas. Secondly, one of the key features of our method is its ability to function on casually captured real-world 360\textdegree{} panoramas, meaning we needed a dataset that meets this criteria. We captured 360\textdegree{} videos of different environments using the Ricoh Theta V. Most of our scenes were captured with a handheld camera. We ensured that our dataset was captured under a variety of environments so that we could evaluate the scalability and robustness of our method. For detail of the captures scene please refer to the appendix.

We sample every 10th frame from the original video for our dataset while avoiding few frames with motion blur caused by casual capturing setup, and this produces around 50cm intervals across input panoramas. 
\section{Results}
\label{sec:RandD}
We compared our results quantitatively and qualitatively with three closely related state-of-the-art methods: Unstructured Lumigraph Rendering (ULR) \cite{buehler2001unstructured}, Inside-Out \cite{hedman2016scalable}, and DeepBlending \cite{hedman2018deep}. They are implemented in the system provided by Bonopera et al. \cite{sibr2020}. We also performed a qualitative comparison with the recent Neural Radiance Field (NeRF) \cite{mildenhall2020nerf} and OmniPhotos\cite{bertel2020omniphotos}. Since NeRF's default projection method only has front facing and out-side in camera poses, we implemented a new projection method for 360\textdegree{} cameras, so we can directly use 360\textdegree{} panorama as input for NeRF. We trained NeRF with 250k iterations on each of the tested scenes. Since the prior works \cite{buehler2001unstructured, hedman2016scalable, hedman2018deep} do not support 360\textdegree{} panoramas directly, the inputs of the above methods are the same as what we used in COLMAP (\autoref{sec.4}), which are 8 perspective images per 360\textdegree{} panorama.

\subsection{Qualitative Results}
\autoref{samplegeneratedview}  shows motion parallax of the synthesis view in various directions. \autoref{eqviews} shows examples of synthesised panoramas, which provide novel panoramic views away from the captured path.

\begin{figure}
\centering
    \includegraphics[width=\linewidth]{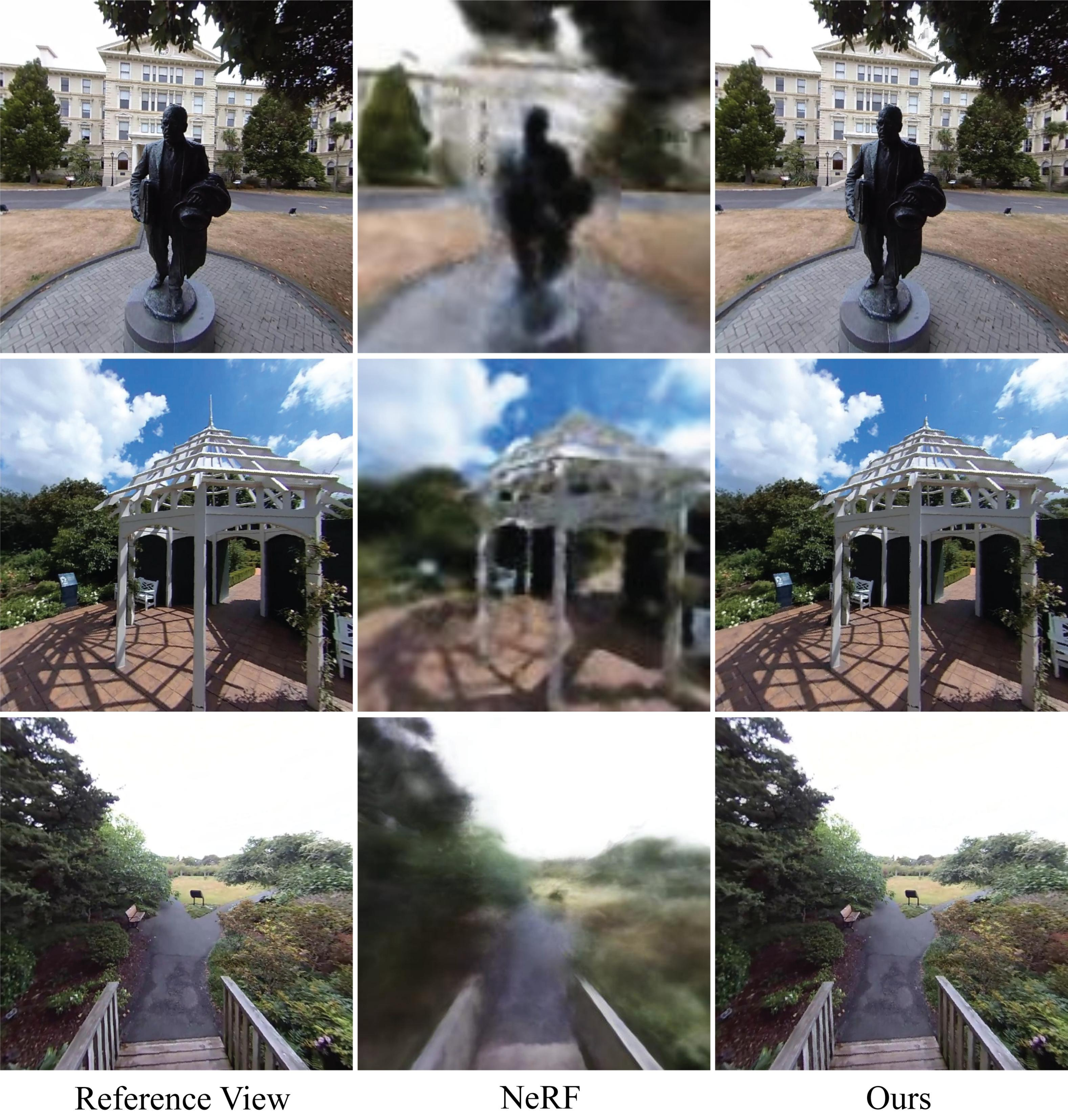}
\caption{Qualitative Comparison with NeRF on different outdoor scenes.}
\label{nerfview}
\end{figure}

Comparisons of the view synthesis of the indoor scenes are in \autoref{qualitative_in}. We noticed that learning-based methods such as DeepBlending efficiently reduce visual artifacts caused by the geometry estimation error, but it often blurs the scene. Both our method and Inside-Out preserve the sharp appearance of the scene, but, compared with Inside-Out, our method shows fewer visual artifacts on boundary regions. 

Comparisons of view synthesis of the outdoor scenes are in \autoref{qualitative_out}, where our method outperforms the prior works. The geometry reconstruction of a large outdoor scene is very challenging. DeepBlending and Inside-Out are highly dependant on the mesh quality by the surface reconstruction. The sparse depths produced by COLMAP tend to contain large missing regions, and therefore may cause surface reconstruction errors. Furthermore, their depth refinement does not consider depth consistency across scenes, and therefore often produces different errors across input panoramas. When blending multiple inputs for novel view synthesis, errors in different scenes are accumulated, producing visual artifacts.

Our depth estimation and refinement method generates reliable and consistent depth information for synthesizing panoramas in different viewpoints. 
Our method recovers detailed geometry with reduced errors and improves results by fusing sparse depths. However, one side effect is that it may remove some of the finer details, causing ghosting artifacts in some cases. Even with this side effect, our method consistently produces higher quality results than prior methods in all our test scenes.

\begin{figure}
\centering
    \includegraphics[width=\linewidth]{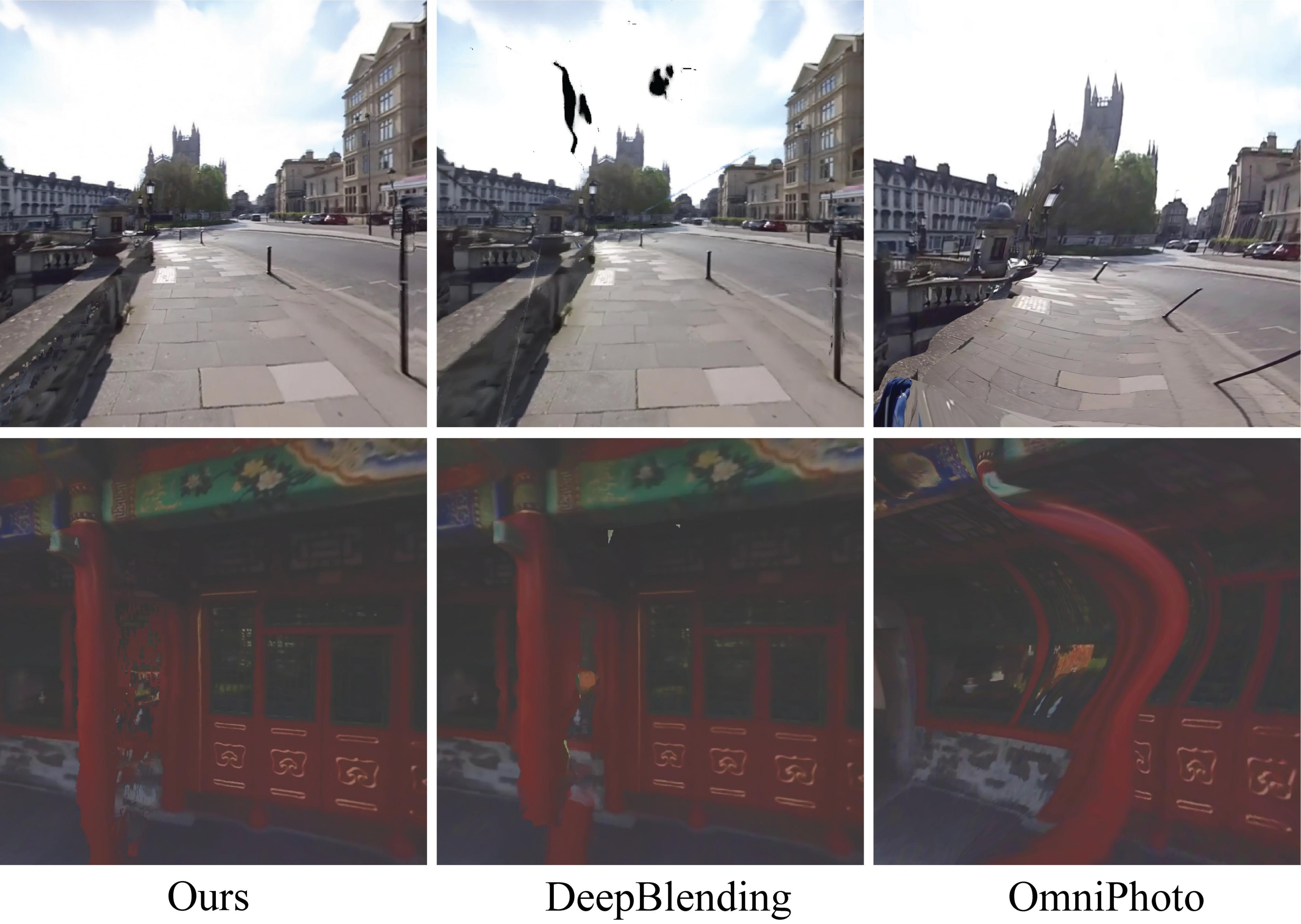}
\caption{Qualitative Comparison with OnmiPhoto on two selected scenes.}
\label{omniphoto_c}
\end{figure}

\begin{figure*}
\centering
    \includegraphics[width=\linewidth]{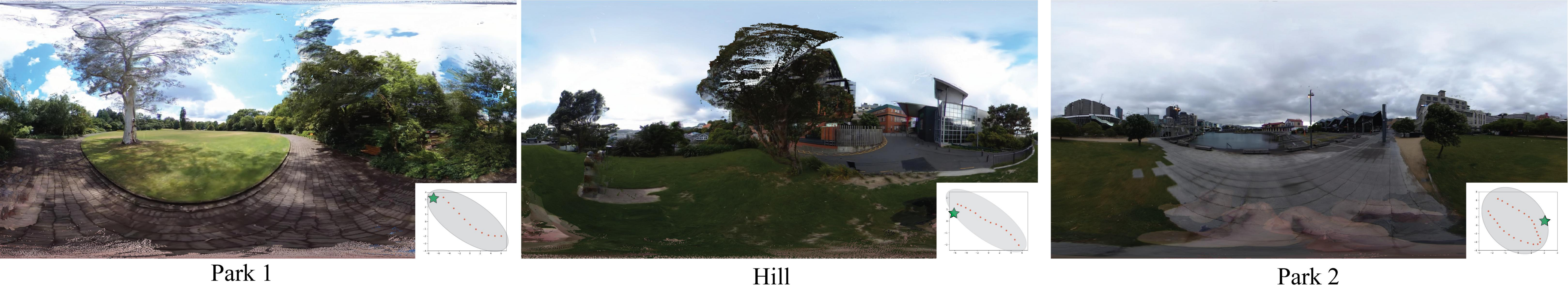}
\caption{Examples of the synthesized view. The \textbf{green star} illustrates the position where the view was synthesized. The \textbf{red point} on the bottom right of each equirectangular image represents the positions (top-down view) of where the input panoramas were captured. The \textbf{gray area} is the potential region for user to move around with a novel synthesized view.}
\label{eqviews}
\end{figure*}

We have tested our method with scene data used in Omniphoto~\cite{bertel2020omniphotos}, and the results are shown in \autoref{omniphoto_c}. Similar to MegaParallax~\cite{bertel2019megaparallax} Omniphoto relies on optical flow to guide image warping. Their estimated optical flow describes the movement of pixels from one image to another, and thus, is only valid within the capture circle. When a user moves outside of the capture circle, optical flow will no longer be able to provide adequate guidance for image warping, and the method has to rely on the estimated geometry proxy. However, their geometry proxy only provides rough geometry information. Using it to perform image warping produces significant visual artifacts, such as twisted objects. Compared to Omniphoto, our method and DeepBlending both can synthesize reasonable results outside the captured regions. In our test, our method shows better visual quality than deep-blending. 

In our test, NeRF struggles to generate novel views using our dataset. We hypothesize that our test scene attempts to cover a very large area with 360\textdegree{} panorama images, the scale of the scene exceeds the capability of NeRF's MLP network, which leads to poor results. We further compared our method with NeRF, and the result is shown in \autoref{nerfview}. 

Previous methods \cite{hedman2016scalable, hedman2018deep} were designed with dense inputs, and therefore we have tested with twice as short intervals. However, their results were not improved in our experiment. We hypothesize that the distortion of 360\textdegree{} panorama produced by the current 360\textdegree{} camera model is bad for convention reconstruction methods, increasing the sample frames introduce more noise features, which do not help improve the quality of the reconstructed mesh.
Similar to previous methods, the number of images also impacts the quality of our method. We demonstrate how our method performs the other methods with the different number of input images in \autoref{diff_inputs}. The increase in the number of input images would improve the result, but would also require more time for it to be processed. Nonetheless as shown in the table, for our method an input of 20 images from a 14 second capture video sequence is sufficient for an acceptable reconstruction result.

\subsection{Quantitative Result}
\label{sec:result}
We quantitatively evaluated our method by comparing it with three closely-related methods~\cite{buehler2001unstructured, hedman2016scalable, hedman2018deep}. The evaluation method we use
follows the approach used by Waechter et al.~\cite{waechter2017virtual}. The idea is to render novel views
images at the same viewpoint as each of the input images, then use the input images as the ground truth to compare with in terms of
completeness and visual quality~\cite{hedman2017casual}.

We tested each method on six scenes with three indoor (PathWay, Hallway, Office) and three outdoor scenes (Bridge, Botanic Garden, Car Park). We did not evaluate the top/bottom view since viewers tend to fixate on the equator when exploring the panorama \cite{bertel2020omniphotos}. We measure the visual quality of the synthesized view at each location quantitatively by comparing the synthesized view with the ground truth image in several metrics such as Multiscale Structural Similarity Index (MS-SSIM), peak signal-to-noise ratio (PSNR), and the perceptual similarity (LPIPS). The average scores of our tests are in Table~\ref{quantitative}. As shown in the table, our method outperforms others in all metrics in both indoor and outdoor scenes.

We also performed a set of ablation studies and show the result in Table \ref{quantitative}. It shows that synthesis results with only sparse depth estimation are the worst, only dense depth estimation improves the results, and a combination of the two shows a better result than each method. Also, the final depth refinement step shows a clear improvement.

\begin{table}[t]
\centering
\begin{tabular}{|l|c|c|c|}
\hline
 & 56 inputs & 33 inputs & 20 inputs \\
\hline
Process-time & $< 18$ hours & $< 8$ hours  & $< 5$ hours  \\
PSNR$\uparrow$ & $23.75$ & $22.05$ & $20.79$ \\ 
SSIM$\uparrow$ & $0.93$ & $0.91$ & $0.87$ \\ 
LPIPS$\downarrow$ & $0.05$ & $0.07$ & $0.10$ \\ 
\hline
\end{tabular}
\caption{The impact of number of input panoramas to the process time and quality.}
\label{diff_inputs}
\end{table}

\begin{table}[t]
\centering
\begin{tabular}{|l|c|c|c|}
\hline
\multicolumn{4}{|c|}{\bfseries Indoor Scenes}\\
\hline
 & PSNR $\uparrow$ &MS-SSIM$\uparrow$ & LPIPS$\downarrow$\\
\hline
ULR & $21.49$ & $0.85$  & $0.13$  \\ 
Inside-Out & $22.51$ & $0.85$ & $0.08$ \\ 
DeepBlending & $22.31$ & $0.82$ & $0.10$ \\ 
\hline
Our method with full process & $\mathbf{25.25}$ & $\mathbf{0.92}$ & $\mathbf{0.07}$ \\ 
\hline
Our with only sparse depth & $14.74$ & $0.68$ & $0.19$ \\ 
Our with only dense depth & $22.46$ & $0.85$ & $0.11$ \\  
Our with only depth fusion & $24.18$ & $0.90$ & $0.09$ \\ 
\hline
\multicolumn{4}{|c|}{\bfseries Outdoor Scenes}\\
\hline
 & PSNR $\uparrow$ &MS-SSIM$\uparrow$ & LPIPS$\downarrow$\\
\hline
ULR & $10.72$ & $0.79$ & $0.53$  \\ 
Inside-Out & $17.3$ & $0.74$ & $0.18$ \\ 
DeepBlending & $16.11$ & $0.71$ & $0.22$ \\ 
\hline
Our method with full process & $\mathbf{21.93}$ & $\mathbf{0.84}$ & $\mathbf{0.12}$ \\ 
\hline
Our with only sparse depth & $11.12$ & $0.50$ & $0.33$ \\ 
Our with only dense depth & $19.98$ & $0.75$ & $0.15$ \\  
Our with only depth fusion & $20.09$ & $0.81$ & $0.14$ \\ 
\hline
\end{tabular}
\caption{Quantitative Comparison of our method with prior works and ablated version of our method. The score are the mean score over all scene indoor/our door scenes. $\uparrow$ means higher is better and $\downarrow$ means lower is better.}
\label{quantitative}
\end{table}

\subsection{Performance}
Our results were tested on a desktop PC with an Intel Xeon W-2133 CPU at 3.60GHz, 16GB of system RAM, and an Nvidia RTX 2080Ti GPU. 
The performance of novel panorama synthesis varies with the number of input panoramas. Throughout the experiment, we have used a set of the closest four input panoramas to synthesize a novel panorama. The data used for the view synthesis method is a set of 4K ($4094\times2048$) equirectangular images and their corresponding 2K ($2048 \times 1024$) depth panoramas. We are able to synthesize the panoramas at 30fps while rendering to the Oculus Rift headset. This frame rate increases to around 90fps when synthesizing a novel panorama at a resolution of $2048 \times 1024$. 

The time it takes for our captured panorama collection to process depends on the number of images, as shown in \autoref{diff_inputs}. We report the time and the amount of storage it requires to process a set of 20 images using our method and prior methods in \autoref{process_time}.

\begin{table}[t]
\centering
\begin{tabular}{|l|c|c|c|}
\hline
 & Deep-Blending & Inside-Out & Ours \\
\hline
COLMAP & $< 3$ hours & $< 3$ hours  & $< 3$ hours  \\
Depth Refinement & $< 1$ hours & $< 1$ hours  & $< 1.5$ hours\\
Storage require & 2.9GB & 3.9GB  &  47MB\\
\hline
\end{tabular}
\caption{The processing time of a scene of 20 input panoramas for the tested methods.}
\label{process_time}
\end{table}

The storage space needed for the final, processed data-set, of 20 source views, is about 47MB. Depth estimation requires a significant space for COLMAP to store their processing data, for instance, the 20 inputs scene required about 3GB of storage. However, once the sparse geometry reconstruction is completed, only the camera information and reconstructed sparse point cloud (sparse depths) are kept, and the rest of data can be discarded.

\section{Discussion and future work}
\label{sec:futureworks}
Our method shows promising results in our tests, outperforming the prior works. However, producing high fidelity 6-DoF video from a single handheld 360\textdegree{} camera is challenging, with a lot of room to improve.
\\

\noindent{\textbf{Depth Estimation with Learning-based methods:}} Our depth estimation could be improved for both better depth accuracy and faster processing. The recent NeRF \cite{mildenhall2020nerf} could provide a clue to improve this. Although, we observed that naive NeRF has converging issues when we train it with 360\textdegree{} images, which we believe the issue is related to the MLP network. The main idea of NeRF is to use a learning-based method to learn the representation of scene. The scene information is still described using similar techniques from the classical MVS, such as ray sampling. When performing ray sampling the images will be converted to 3D points, thus, the format of the image does not effect the result of this process. Furthermore, as 360\textdegree{} images have advantages of FoV compared to normal perspective images, the use of 360\textdegree{} images should improve the result. Thus, replacing the MLP network in NeRF with a more effective learning structure, and use the improved NeRF for depth estimation, could be a good next step to improve our pipeline.
\\

\noindent{\textbf{Reflective and Transparent Surface:}}
Our depth estimation has limitations for reflective and transparent surfaces such as glass or water. An example of these limitations are shown in \autoref{eqviews} (Hill scene), where our method contains ghosting artifacts on the glass window. The Multi-Plane Image (MPI) representation has demonstrated its ability to reproduce the behaviour of reflective and transparent surfaces \cite{zhou2018stereo, flynn2019deepview, broxton2020immersive}. Adapting similar methods into our pipeline to separate the transparent layers would be an interesting next step. 
\\

\noindent{\textbf{Dynamic Objects:}}
Similar to previous work~\cite{hedman2016scalable, choi2019extreme, hedman2018deep, mildenhall2020nerf}, our method has limitations with dynamic objects. Given that the input of our method is a capture from a single camera, moving objects will shown in different positions in each image, making it difficult to estimate consistent depth for the object. Furthermore, the different positions also cause inconsistent projection for the object (same object projected into a different place). This can be solve by estimated the flow of the dynamic object over time~\cite{park2021nerfies, park2021hypernerf}, and use flow information to guide image synthesis.
\\

\begin{figure}
\centering
    \includegraphics[width=\linewidth]{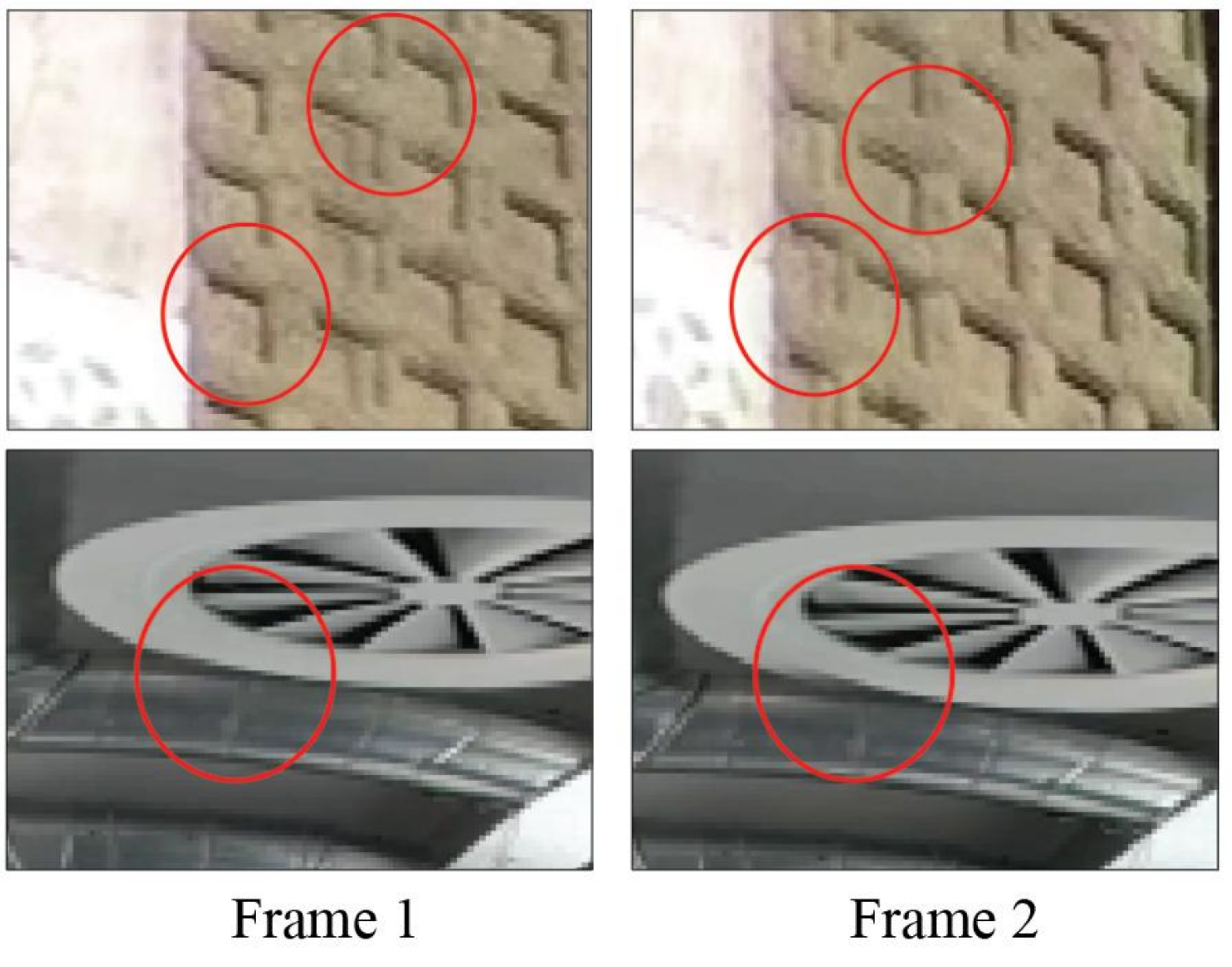}
\caption{Examples of the stitching artifact. The red circle shows where the image stitching artefact appears.}
\label{stich_ex}
\end{figure}

\noindent{\textbf{Stitching artifact:}}
360\textdegree{} panoramas are produced by multi-image stitching, which with current methods still contains small distortions or gaps around the image boundaries, causing inconsistencies of object shape in these areas. We show an example of such artifacts in \autoref{stich_ex}. This causes ghosting artifact in the view synthesis when performing image blending. Reducing stitching artifacts in the captured 360\textdegree{} panoramas could reduce such artifacts. 
\\

\noindent{\textbf{Pre-processing:}}
Our method relies on off-line pre-processing including depth estimation and refinement steps. It is acceptable for a static scene, but limited when adapting dynamic scenes. 
\\

\noindent{\textbf{User-study:}}
Overall user studies are highly restricted due to COVID-19 at the time of study; nationwide lockdown restricting any mass gathering. Whenever available, we plan to perform user study to measure presence, immersion, VR sickness, and the overall usability of 6-DoF experiences produced by our methods.

\section{Conclusion}
We present a novel and complete pipeline to synthesize a novel view panorama using a set of 360\textdegree{} images captured by a handheld 360\textdegree{} camera. The result provides 6-DoF experiences, a sensation to walk around a captured large scale scene with support of full panoramic view in any location. We have tested our method in various test scenes captured by a single handheld 360\textdegree{} camera including indoor and outdoor scenes, and mid to large scale scenes. We also compared our results with current state-of-the art methods, showing better visual quality and robustness. Our method consistently produces high fidelity results across all test scenes, while previous method sometimes failed. We also outline the current limitations and potential future work that could lead to improvements. We believe our method, tested on a consumer graded 360\textdegree{} camera, will be easily adapted to various applications that requires 6-DoF experiences of a captured large scale scenes, particularly benefiting casual users.


%




\ifCLASSOPTIONcaptionsoff
  \newpage
\fi



%
\bibliographystyle{abbrv-doi}
\bibliography{sample-base}



%








\end{document}